\pdfoutput=1

\documentclass[11pt]{article}

\usepackage{acl}

\usepackage{times}
\usepackage{latexsym}

\usepackage[T1]{fontenc}

\usepackage[utf8]{inputenc}

\usepackage{microtype}

\usepackage{array}
\usepackage{multirow}
\usepackage{makecell}
\usepackage{soul}
\usepackage{url}
\usepackage{graphicx}
\usepackage{amsmath}
\usepackage{booktabs}
\usepackage{subfigure}
\usepackage{algorithmic}
\usepackage{enumitem}
\usepackage{amsthm,amsmath,amssymb}
\usepackage{mathrsfs}
\usepackage{bm}
\usepackage{hyperref}
\usepackage[ruled,linesnumbered]{algorithm2e}

\title{Semantic-Preserving Adversarial Code Comprehension}

\author{Yiyang Li$^{1,2}$, Hongqiu Wu$^{1,2}$ \and Hai Zhao$^{1,2,}$\thanks{\; Corresponding author. This paper was partially supported by Key Projects of National Natural Science Foundation of China (U1836222 and 61733011).} \\
        $^1$ Department of Computer Science and Engineering, Shanghai Jiao Tong University\\
        $^2$ Key Laboratory of Shanghai Education Commission for Intelligent Interaction\\and Cognitive Engineering, Shanghai Jiao Tong University\\
        \texttt{\{eric-lee,wuhongqiu\}@sjtu.edu.cn,zhaohai@cs.sjtu.edu.cn}
}

\begin{document}
\maketitle

\begin{abstract}
Based on the tremendous success of pre-trained language models (PrLMs) for source code comprehension tasks, current literature studies either ways to further improve the performance (generalization) of PrLMs, or their robustness against adversarial attacks. However, they have to compromise on the trade-off between the two aspects and none of them consider improving both sides in an effective and practical way. To fill this gap, we propose Semantic-Preserving Adversarial Code Embeddings (SPACE) to find the worst-case semantic-preserving attacks while forcing the model to predict the correct labels under these worst cases. Experiments and analysis demonstrate that SPACE can stay robust against state-of-the-art attacks while boosting the performance of PrLMs for code${}^1$.
\end{abstract}

\let\thefootnote\relax\footnotetext{${}^1$Our codes and data are available at \url{https://github.com/EricLee8/SPACE}}

\section{Introduction}
Inspired by the tremendous success of large pre-trained language models (PrLMs) in natural language understanding (NLU) field \cite{devlin-etal-2019-bert, liu2019roberta, xlnet, electra}, many attempts have been made to pre-train a language model for source code comprehension, aiming at learning good code embeddings to help code-related downstream tasks like detecting vulnerable codes, searching codes with natural language queries or answering questions about codes \cite{defect, CodeSearchNet, codeqa, cubert, C-BERT, feng-etal-2020-codebert, GraphCodeBert}. Based on this, current literature focuses on either methods to further improve the performance of PrLMs \cite{contrastivecode, contra-stru-func, cotext, wu-etal-2021-code}, or their robustness against adversarial attacks \cite{advref1aaai, advref2, natural-attack, obfuscations}. However, they have to compromise on the trade-off between the two aspects and none of them consider improving both sides in an effective and practical way \cite{advorbuts-tradeoff1, tradeoff2}. For example, CodeBERT \cite{feng-etal-2020-codebert} and GraphCodeBERT \cite{GraphCodeBert} are two most widely used PrLMs for code and they have significantly boosted the performance of source code comprehension models. Unfortunately, these powerful models are vulnerable to adversarial attacks that slight perturbations can make them produce wrong labels \cite{natural-attack}. To tackle this problem, previous works attempt to adopt adversarial training framework or data augmentation tricks to improve model robustness, yet they fail to maintain accuracy on the origin clean test set \cite{advorbuts-tradeoff1, tradeoff2}, or are too complicated to be practical \cite{semanticroust}.

A natural question can be asked here: is there a simple yet effective way to subtly avoid the trade-off between performance and robustness and make them both improved? To make the answer yes, we propose Semantic-Preserving Adversarial Code Embeddings (SPACE) to find the worst-case semantic-preserving attacks while forcing the model to predict the correct labels under these worst cases. Different from previous works which also adopt adversarial training on codes \cite{semanticroust, advorbuts-tradeoff1}, we search for the inner worst-case perturbations in the continuous embedding space instead of the discrete token space. Compared with the latter, which is usually a complicated combinational optimization problem, our method is naturally differentiable which makes it easy to be incorporated into the gradient-based training framework and much more efficient. Besides, adding perturbations to the embedding layer is a natural solution to avoid the aforementioned trade-off between performance and robustness, which is observed and proved on multiple NLU tasks on plain texts \cite{AdvTextClassification, robustMT}. Nonetheless, its efficacy on source code comprehension tasks has not been fully investigated. Besides, our experiments show that simply adopting vanilla adversarial training is not enough to gain significant improvements in the context of programming languages, since it can not preserve the semantic meaning and grammatical correctness.

Therefore, in this paper, we combine adversarial training with the data characteristics of programming languages to propose SPACE: we find ways to generate worst-case perturbations that preserve the semantic meaning and grammatical correctness of programming languages. Experiments show that our SPACE is superior to the vanilla adversarial training and significantly outperforms strong baselines of CodeBERT and GraphCodeBERT on several datasets while staying robust against the SOTA attacking methods.

To sum up, the contributions of our SPACE are the following four folds:
\begin{itemize}[leftmargin=*, topsep=1pt]
    \setlength{\itemsep}{0pt}
    \setlength{\parsep}{0pt}
    \setlength{\parskip}{0pt}
    \item To the best of our knowledge, we are the first to explore the efficacy of adversarial training on the continuous embedding space for source code comprehension tasks.
    \item We innovatively combine adversarial training with the data characteristics of programming languages to propose SPACE, which is able to boost the performance and robustness of PrLMs for code simultaneously.
    \item SPACE is practical and effective, which is naturally differentiable and easy to be incorporated into the gradient-based training framework with only a few lines of code.
    \item Experiments and analyses on several benchmark datasets demonstrate that our SPACE outperforms strong baselines by large margins, and stays robust against state-of-the-art attacks.
\end{itemize}

\section{Related Work}
\subsection{Pre-trained Language Models for Code}
In recent years, the prosperity of PrLMs has been witnessed in natural language understanding field. Inspired by this, researchers start to investigate the power of PrLMs for source code comprehension tasks. To name a few, \citet{Pre-trainedContextual} adopt masked language modeling (MLM) and next sentence prediction (NSP) to pre-train BERT on Python corpus. \citet{C-BERT} propose C-BERT, which makes use of a carefully designed tokenizer and whole word masking (WWM) objective to pre-train on C language. \citet{feng-etal-2020-codebert} present CodeBERT, a bi-modal PrLM trained on massive code-text pairs using masked language modeling (MLM, \citealt{devlin-etal-2019-bert}) and replaced token detection (RTD, \citealt{electra}) objectives, in an attempt to handle bi-modal tasks like natural language code search. \citet{GraphCodeBert} design GraphCodeBERT, an even stronger PrLM that leverages the graph-structured code data flow to model the dependency relation between variables. There are many other PrLMs for source code \cite{contrastivecode,Corder,wang2021syncobert}. We do not try our SPACE on all of them since it will cost too much time and computational resources. Without loss of generality, in this paper, we choose CodeBERT and GraphCodeBERT, two most widely-used and easy-to-use models, as our backbone.

\begin{figure*}[tbp]
	\includegraphics[width=1.0\textwidth]{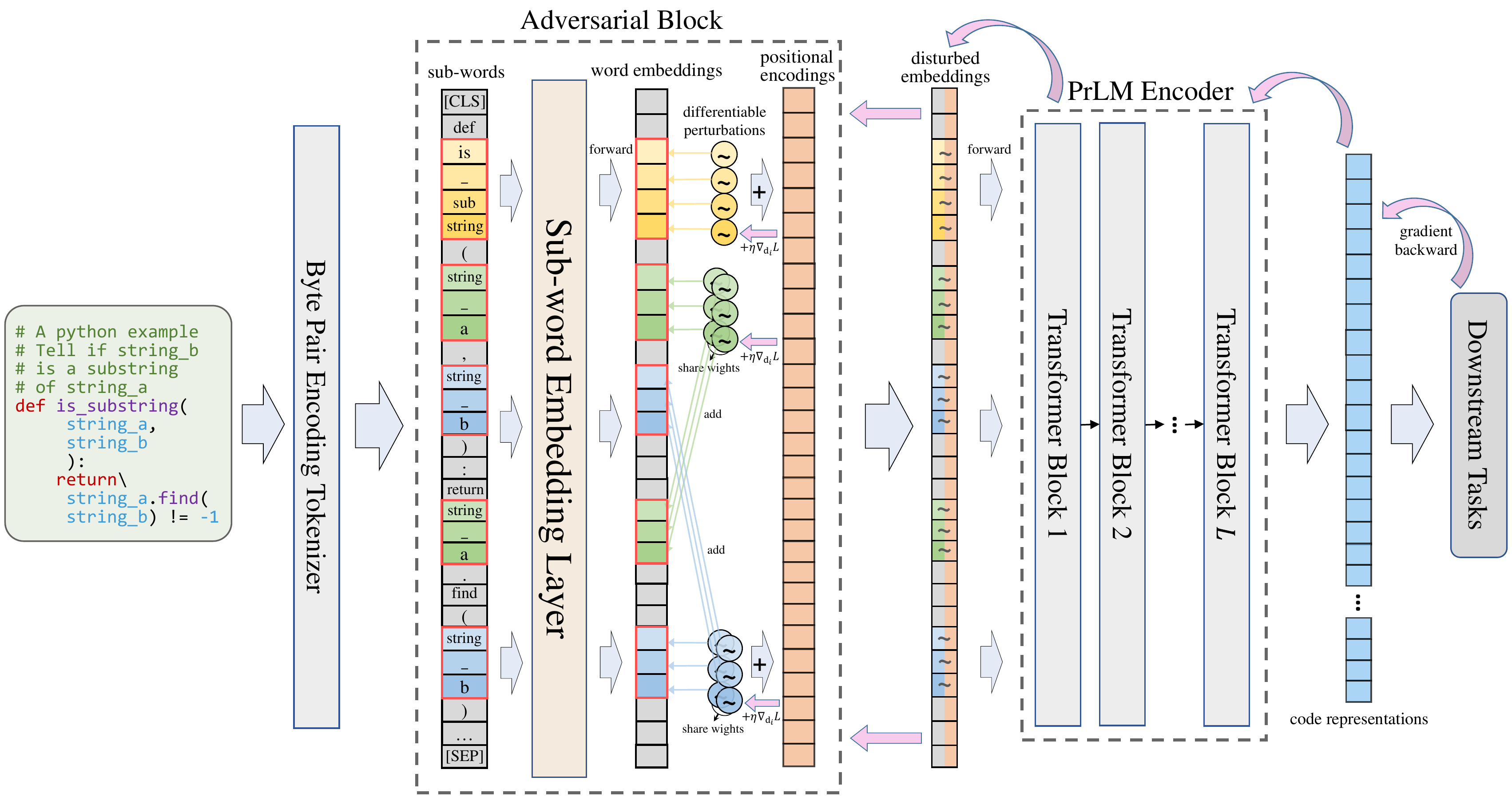}
	\centering
	\caption{The overview of our model, which consists of three parts: a Byte Pair Encoding (BPE) Tokenizer, an Adversarial Block and a PrLM Encoder. The detailed process is explained in Section \ref{sec:adversarial_process}.}
	\label{fig:overview}
\end{figure*}

\subsection{Attacks and defenses on Code}
Aiming at testing and improving the robustness of code models, researchers have conducted a series of experiments on attacks and defenses.

For attacks, \citet{obfuscations} utilize program obfuscations, which are conventionally used to avoid attempts at reverse engineering programs, as adversarial perturbations. They design alternate optimization (AO) and joint optimization (JO) to solve the problem of where and how to inject obfuscations, which yield a satisfying attack success rate. \citet{advref1aaai} propose a Metropolis-Hastings sampling-based identifier renaming technique (MHM) to efficiently generate semantic-preserving adversarial examples. \citet{natural-attack} present ALERT (Naturalness Aware Attack) that considers the natural semantic of generated adversarial examples. They make use of the masked language prediction function of PrLMs to produce a ranked list of potential substitutes for each token, and design genetic algorithms to search for the optimal combination of variables and corresponding substitutes. ALERT is the state-of-the-art attack method and the first to attack PrLMs, yet our experiments demonstrate that our SPACE is able to maintain robustness under ALERT by reducing the attack success rate over 10\%.

For defenses, \citet{semanticroust} adopt adversarial training framework to improve the robustness of models against various code transformations (e.g. dead code insertion and identifier renaming,). They define the inner maximization problem of adversarial training as finding the worst code transformations, which is a discrete combinational optimization problem. To solve it, they limit the number transformations to $1$ and deploy a complicated program sketching trick to approximately solve the inner maximization in a gradient-based way. Their work is too complicated to be practical, on the contrary, our SPACE is performed on the continuous embedding space, which is naturally compatible with the gradient-based training framework and easy to apply with only a few lines of code. \citet{advorbuts-tradeoff1} also employ adversarial training framework specifically for the type prediction task. They design uncertainty scores and a program abstractor to solve the inner combinational optimization problem using integer linear programming (ILP). Their work suffers the trade-off between robustness and performance, while our SPACE is capable of improving both aspects simultaneously.

\section{Methodology}
In this section, we will first briefly introduce the background knowledge: the general adversarial training and it on the continuous embedding space, then present the proposed SPACE in detail.

\subsection{Background: Adversarial Training}
Adversarial training is defined as a min-max problem, where a model should be learned to minimize the maximum risk posed by the attacking perturbations \cite{advori}. Formally, it can be formulated as follows:
\begin{equation}
    \min \limits_{\bm{\theta}} \mathbb{E}_{(\mathbb{X}, y)\sim \mathcal{D}}\left[\max \mathcal{L}(f_{\bm{\theta}}(\mathcal{A}(\mathbb{X})), y)\right]
\end{equation}
Here $\mathbb{X}$ and $y$ are the input and its corresponding label sampled from data distribution $\mathcal{D}$, $\mathcal{L}$ is the loss function, $f_{\bm{\theta}}$ is the target model with trainable parameter $\bm{\theta}$ and $\mathcal{A}$ is the adversarial transformation applied to $\mathbb{X}$.

Previous works that adopt adversarial training on source code comprehension tasks usually define $\mathcal{A}$ as some semantic-preserving code transformations, which yields \emph{actual} adversarial examples in the discrete token space and is hard to optimize \cite{advorbuts-tradeoff1, semanticroust}. Different from them, we propose to define $\mathcal{A}$ as perturbations on the continuous embedding space:
\begin{equation}
    \label{eq:myadv}
    \min \limits_{\bm{\theta}} \mathbb{E}_{(\mathbb{X}, y)\sim \mathcal{D}}\left[\max \limits_{\Vert\bm{\delta} \Vert_F \le \epsilon} \mathcal{L}(f_{\bm{\theta}}(\mathcal{A}(\mathbb{X}, \bm{\delta})), y)\right]
\end{equation}
where $\bm{\delta}$ is the perturbation matrix on $\mathbb{X}$, whose Frobenius norm is constrained to be less than $\epsilon$. This definition produces \emph{virtual} adversarial examples in the embedding space, which is easy to optimize via gradient-based methods.

\begin{algorithm}[tb]
    \small
    \caption{Semantic-Preserving Adversarial Training for Programming Languages}
    \label{alg:SPACE}
    \KwData{Training samples $\mathcal{D}$ = \{($\mathbb{X}, y$)\}, perturbation bound $\epsilon$, learning rate $\alpha$, ascent steps $K$, adversarial learning rate $\eta$.}
    \KwOut{Model parameters $\bm{\theta}$.}
    Initialize model parameters $\bm{\theta}$\\
    \For{$\operatorname{epoch}=1,2,\dots,N_{ep}$}{
        \For{$\operatorname{mini-batch} B \subset \mathcal{D}$}{
            \For{$i=1,2,\dots,m$}{
                $\bm{\delta}_0^i \leftarrow \bm{0}\in \mathcal{R}^{k_i\times d}$
            }
            $g_0\leftarrow \bm{0}$\\
            \For{$t=1,2,\dots, K$}{
                $\bm{\delta}_{t-1}\leftarrow \bm{0}\in \mathcal{R}^{n\times d}$\\
                \For{$p=1,2,\dots,n$}{
                    \uIf{position $p$ is the $j_{\operatorname{th}}$ sub-word of identifier $d_i$}{
                        $\bm{\delta}_{t-1}[p]\leftarrow \bm{\delta}_{t-1}^i[j]$
                    }
                }
                $g_t\leftarrow g_{t-1} + \frac{1}{K}\mathbb{E}_{(\mathbb{X},y)\in B}[\nabla_{\bm{\theta}} \mathcal{L}( f_{\bm{\theta}}(\mathcal{A}(\mathbb{X}, \bm{\delta}_{t-1})), y )]$\\
                \For{$i=1,2,\dots,m$}{
                    $g_{adv}^i\leftarrow \nabla_{\bm{\delta}_{t-1}^i} \mathcal{L}(f_{\bm{\theta}}(\mathcal{A}(\mathbb{X}, \bm{\delta}_{t-1})), y)$\\
                    $\bm{\delta}_t^i\leftarrow \Pi_{\Vert \bm{\delta}_{t-1}^i \Vert_F \le \epsilon} (\bm{\delta}_{t-1}^i + \eta\cdot g_{adv}^i / \Vert g_{adv}^i \Vert_F)$
                }
            }
            $\bm{\theta}\leftarrow \bm{\theta} - g_K$
        }
    }
    return $\bm{\theta}$
\end{algorithm}

\subsection{Adversarial Training on Embeddings}
Before presenting SPACE, we first introduce the vanilla adversarial training on embedding space.

Given an input code sequence, we first tokenize it into multiple sub-words using a Byte Pair Encoding Tokenizer (BPE, \citealt{BPE}), which is illustrated in the left part of Figure \ref{fig:overview}. After obtaining the sub-word sequence $\mathbb{X} = \{w_i\}_{i=1}^n$, we embeds them into a high-dimensional embedding space with an embedding look-up table $\phi$: $E = \phi(\mathbb{X}) = \{e_i\}_{i=1}^n \in \mathcal{R}^{n\times d}$, where $d$ is the dimension of the embedding space. At this point, we are supposed to find the optimal $\bm{\delta}\in \mathcal{R}^{n\times d}$ to maximize the inner part of Eq. (\ref{eq:myadv}), where $\mathcal{A}(\mathbb{X}, \bm{\delta}) = \phi(\mathbb{X}) + \bm{\delta}$ and $\Vert\bm{\delta} \Vert_F \le \epsilon$. This optimization is non-concave for neural networks, hence we suppose the loss function is locally linear and apply multiple gradient ascent steps to solve it:
\begin{equation}
    \label{eq:pgd}
    \begin{aligned}
        g(\bm{\delta}_t) &= \nabla_{\bm{\delta}_t} \mathcal{L}(f_{\bm{\theta}}(\mathcal{A}(\mathbb{X}, \bm{\delta}_t)), y) \\
        \bm{\delta}_{t+1} &= \Pi_{\Vert \bm{\delta} \Vert_F \le \epsilon} (\bm{\delta_t} + \eta g(\bm{\delta}_t)/\Vert g(\bm{\delta}_t) \Vert_F)
    \end{aligned}
\end{equation}
Here $\Pi$ is a projecting function that constrain $\bm{\delta}$ within the $\epsilon$-ball. It is proved in literature that the above Projected Gradient Descent (PGD) algorithm is able to neatly solve the maximization problem \cite{PGD}.

In our implementation, we utilize the Free Large-Batch algorithm (FreeLB, \citealt{FreeLB}) to improve the efficiency during training. The target model $f_{\bm{\theta}}$ is the PrLM Encoder (see the right part of Figure \ref{fig:overview}), which is CodeBERT or GraphCodeBERT in our experiments.

\subsection{Semantic-Preserving Adversarial Training for Programming Languages}
\label{sec:adversarial_process}
Vanilla adversarial training suffices to produce good virtual adversarial examples for natural language. However, it is not suitable for programming languages, which are grammatically and syntactically strict. The key problems here are two-fold. First, the keywords in programming languages are so sensitive that small perturbations on them could change the functionality of the original code snippet, resulting in label flipping \cite{interpretable}. Second, different perturbations added to instances of the same identifier will lead to different embeddings for the same identifier, which breaks the grammar rule of programming languages. To avoid the first problem, the simplest and most efficient way is to add adversarial perturbations only on the embeddings of identifiers (e.g. names of functions, variables, and classes) and leave the keywords unchanged. For the second problem, we maintain the same adversarial perturbation for each identifier, making sure the embeddings of the same identifier are still identical after perturbing. In conclusion, we aim to find constrained perturbations that preserve the semantic meaning and grammatical correctness of programming languages.

Motivated by the above discussion, we propose Semantic-Preserving Adversarial Code Embeddings (SPACE). Specifically, for each identifier $d_i$ that contains $k_i$ sub-words $d_i = \{s_{ij}\}_{j=1}^{k_i}$, we maintain a specific perturbation $\bm{\delta}_i\in \mathcal{R}^{k_i\times d}$ for it. Each perturbation $\bm{\delta}_i$ is then copied and assigned to the positions of its corresponding identifier instances in the code snippet. As illustrated in the Adversarial Block of Figure \ref{fig:overview}, the BPE Tokenizer tokenizes the input code sequence into sub-word tokens, which are then embedded into high-dimensional word embeddings by a Sub-word Embedding Layer. After that, we maintain the differentiable perturbations for each sub-word that belongs to the same identifier, and add them to the corresponding word embeddings and positional embeddings to form the distributed embeddings, which are finally further encoded by the PrLM Encoder. In Figure \ref{fig:overview}, different colors represent different identifiers and their shade indicates different sub-words. The differentiable perturbations are updated according to their gradients to the final loss using the gradient ascent algorithm, in order to maximize the negative impact of the perturbations to the model.

The whole training process is shown in Alg. (\ref{alg:SPACE}). We first initialize the model parameters $\bm{\theta}$, then for every mini-batch in an epoch, we initialize the perturbations for all identifiers, $\bm{\delta}^i_0$, to zero. After that, $K$ steps of adversarial training will be conducted. In each step $t$, the perturbations of each identifier $\bm{\delta}^i_{t-1}$ will be assigned to the positions of their corresponding identifier instances to form $\bm{\delta}_{t-1}$, which is then used to attack the model in order to accumulate the gradient of $\bm{\theta}$ and update each $\bm{\delta}^i_t$ for the next step. After $K$ steps, we update $\bm{\theta}$ using the accumulated gradient $g_K$. Experimental results and analysis in the following sections demonstrate that the obtained model is better than normal training and stays robust against adversarial attacks.

\section{Experiments}
To comprehensively evaluate our SPACE, we conduct experiments on three types of source code comprehension tasks: code defect detection task for classification style, natural language code search task for retrieval style, and code question answering task for generative style. Meanwhile, they are code-to-code, natural-language-to-code and code-to-natural-language tasks, respectively. In this section, we will first introduce each task and its corresponding dataset, then present the results of them.

\subsection{Tasks and Datasets}
\textbf{Code Defect Detection} requires a model to predict whether a code snippet contains malicious parts that may attack software systems. We conduct this experiment on the Defects4J dataset \cite{defect}, which defines defect detection as a binary classification task.\\
\textbf{Natural Language Code Search} is to retrieve the most relevant code snippet from a database given a natural language query, which is helpful to reuse codes when developing new systems. We conduct this experiments on the CodeSearchNet dataset \cite{CodeSearchNet} and follow \citet{GraphCodeBert} to process the data. This dataset contains six subsets for different languages: Ruby, JavaScript, Go, Python, Java, and PHP.\\
\textbf{Code Question Answering} aims at learning a model to answer questions with regard to a snippet of code. In this setting, the answer should be generated as natural language by a decoder rather than copied from the source code. We conduct this experiment on CodeQA dataset \cite{codeqa}, which contains two subsets for Java and Python.

For statistics about these datasets, please refer to Appendix \ref{app:datasetstat}. For more details about our experimental settings (e.g. hyper-parameters, environments, and hard-wares), please refer to Appendix \ref{app:expsetting}.

\subsection{Experimental Results}
\label{sec:expresults}

\begin{table}[tbp]
    \centering
    \small
    \begin{tabular}{l c c}
        \specialrule{0.09em}{0.0pt}{1.8pt}
        \multirow{2}{*}{\textbf{Model}} & \multicolumn{2}{c} {\textbf{Accuracy}} \\
        & \small{CodeBERT} & \small{GraphCodeBERT}\\
        \specialrule{0.03em}{1.3pt}{0.5pt}
        \specialrule{0.03em}{0.5pt}{1.3pt}
        baseline & 62.08 & 64.06 \\
        \quad +ADV. & 64.85 & 64.45 \\
        \quad +SPACE & \textbf{66.01} & \textbf{66.12} \\
        \specialrule{0.04em}{1.3pt}{1.3pt}
        \quad +augmentation & 63.25 & 63.47 \\
        \quad +rand. ADV. & 64.02 & 64.41 \\
        \quad +rand. SPACE & 64.57 & 65.08 \\
        \specialrule{0.09em}{1.2pt}{0.0pt}
    \end{tabular}
    \caption{Results on Defects4J dataset, where the upper half presents the main results based on two PrLMs and the lower half presents ablation results.}
    \label{tab:defect}
\end{table}

\begin{table*}[tbp]
    \centering
    \small
    \begin{tabular}{l|c|c c c c c c c}
		\specialrule{0.09em}{0.0pt}{0.1pt}
		\multirow{2}{*}{Model}&AdvTest&\multicolumn{7}{c}{CodeSearch}\\
		\cline{2-9}
		& Python & Ruby & JavaScript & Go & Python & Java & PHP & Average \\
		\hline
		CodeBERT & 27.2 & 67.9 & 62.0 & 88.2 & 67.2 & 67.6 & 62.8 & 69.3\\
		\quad +ADV. & 28.3 & 68.5 & 62.3 & 88.5 & 67.5 & 67.6 & 62.6 & 69.5($\uparrow$ 0.2)\\
		\quad +SPACE & \textbf{32.6} & \textbf{69.4} & \textbf{62.7} & \textbf{89.1} & \textbf{68.3} & \textbf{68.5} & \textbf{63.3} & \textbf{70.2} ($\bm{\uparrow}$ \textbf{0.9})\\
		\hline
		\hline
		GraphCodeBERT & 35.2 & 70.3 & 64.4 & 89.7 & 69.2 & 69.1 & 64.9 & 71.3\\
		\quad +ADV. & 37.8 & 70.7 & 64.9 & 89.7 & 69.5 & 69.4 & 64.7 & 71.5 ($\uparrow$ 0.2)\\
		\quad +SPACE & \textbf{41.3} & \textbf{72.1} & \textbf{65.8} & \textbf{90.1} & \textbf{70.6} & \textbf{70.3} & \textbf{65.4} & \textbf{72.4} ($\bm{\uparrow}$ \textbf{1.1})\\
		\specialrule{0.09em}{0.1pt}{0.0pt}
	\end{tabular}
    \caption{Results on CodeSearchNet dataset.}
    \label{tab:nlcodesearch}
\end{table*}

\begin{table*}
    \centering
    \small
    \begin{tabular}{l|ccccc|ccccc}
        \specialrule{0.09em}{0.0pt}{0.2pt}
        \multirow{2}{*}{Model} & \multicolumn{5}{c|}{Python} & \multicolumn{5}{c}{Java} \\
        \cline{2-11}
        & BLEU & ROUGE & METEOR & EM & F1 & BLEU & ROUGE & METEOR & EM & F1 \\
        \hline
        CodeBERT & 34.86 & 30.28 & 12.51 & 4.93 & 31.56 & 32.40 & 28.22 & 10.10 & 6.20 & 29.20\\
        \quad +ADV. & 36.15 & 32.84 & 14.07 & 5.84 & 34.08 & 33.03 & 29.13 & 10.60 & 6.24 & 30.02\\
        \quad +SPACE & \textbf{36.47} & \textbf{33.03} & \textbf{14.14} & \textbf{6.14} & \textbf{34.38} & \textbf{33.49} & \textbf{29.64} & \textbf{11.01} & \textbf{6.42} & \textbf{30.66}\\
        \specialrule{0.03em}{0.0pt}{0.8pt}
        \specialrule{0.03em}{0.8pt}{0.0pt}
        GraphCodeBERT & 35.95 & 32.31 & 13.39 & 5.86 & 33.52 & 33.22 & 29.24 & 10.78 & 6.52 & 30.22\\
        \quad +ADV. & 36.57 & 33.29 & 14.15 & 6.24 & 34.52 & 33.84 & 30.41 & 11.33 & 6.76 & 31.42\\
        \quad +SPACE & \textbf{37.03} & \textbf{33.97} & \textbf{14.65} & \textbf{6.41} & \textbf{35.25} & \textbf{34.11} & \textbf{30.96} & \textbf{11.68} & \textbf{6.92} & \textbf{32.01}\\
        \specialrule{0.09em}{0.2pt}{0.0pt}
    \end{tabular}
    \caption{Results on CodeQA dataset.}
    \label{tab:codeqa}
\end{table*}

\paragraph{Results on Defect Detection} are shown in Table \ref{tab:defect}, we follow \citet{codexglue} to use accuracy score as our evaluation metric. In this section, we focus on the upper half of Table \ref{tab:defect}, where the main results are presented. The baselines here are pure CodeBERT and GraphCodeBERT with a linear classifier on top and +ADV. is the abbreviation for the vanilla adversarial training method. We see from the table that vanilla adversarial training can improve the performance to a certain extent, while our SPACE is constantly better than it and outperforms strong baselines by large margins: over \textbf{3.93\%} and \textbf{1.96\%} on CodeBERT and GraphCodeBERT, respectively. This observation demonstrates that our SPACE, which takes the advantage of source code characteristics, is superior to vanilla adversarial training on all tokens.

\paragraph{Results on Natural Language Code Search} are tabulated in Table \ref{tab:nlcodesearch}, where the evaluation metric is Mean Reciprocal Rank (MRR) following \citet{CodeSearchNet}. We see from the table that our SPACE outperforms vanilla adversarial training and baselines in all languages, demonstrating the universality of our method, which is language-agnostic and can be used to boost the performance of models for different programming languages. Beside, \citet{codexglue} provide an adversarial test set for Python, which replaces all arguments of a function with $\{arg\_0, arg\_1, \dots, arg\_n\}$. We also test our method in this adversarial test set to preliminarily verify the robustness of SPACE. In this setting, we observe slight improvements on robustness using vanilla adversarial training, yet they are still far from satisfactory. In contrast, our SPACE boosts the performance on this test set by \textbf{5.4\%} and \textbf{6.1\%} over the baseline of CodeBERT and GraphCodeBERT, respectively. It is worth noting that the calculation of MRR score in this adversarial test set is different from the clean test set, hence the results are not directly comparable with the original ones. For the robustness test with stronger adversarial attacks, we present it in Section \ref{sec:robusttest}.

\paragraph{Results on Code Question Answering} are shown in Table \ref{tab:codeqa}, where we use PrLMs (CodeBERT and GraphCodeBERT) as the encoder to build representations for the source code and its corresponding question, then apply a Transformer Decoder \cite{VaswaniAttention} to generate answers. Following \citet{codeqa}, we use BLEU, ROUGE-L, METEOR, EM, and F1 as the evaluation metrics. BLEU computes the n-gram overlapping between candidates and references, while METEOR is an improvement of BLEU that considers some grammatical and lexical information, such as synonyms, roots, and affixes. ROUGE-L computes the longest common subsequence (LCS) between the candidates and references, then calculates the F1 ratio by measuring the recall over references and precision over candidates. EM stands for Exactly Match and F1 is the character-wise F1 score between candidates and references. Similar observations can be seen in Table \ref{tab:codeqa} that SPACE yields the best result compared with baselines and vanilla adversarial training for both Python and Java dataset, which further demonstrates the effectiveness of our method in generative tasks.

\section{Analysis}
\label{sec:analysis}
In this section, we conduct a series of experiments and analyses to get a more in-depth understanding of SPACE and its robustness against various adversarial attacks, including state-of-the-art ones.

\subsection{Ablation Study}
There are multiple factors that could be the potential reasons for the performance improvements: more virtual augmented samples by the FreeLB algorithm, or simply the virtual perturbations on embedding space. To investigate whether these factors could yield performance improvements as good as SPACE, we conduct additional experiments on each of these factors.

Notice that in Alg. (\ref{alg:SPACE}), we accumulate the gradient of $\bm{\theta}$ in each of the $K$ steps, resulting in $K$ times more \emph{virtual augmented samples} compared with the original training samples. To simulate the augmented samples and also to explore whether \emph{actual} data augmentation would be enough to boost the performance, we perform semantic-preserving code transformations for each training sample and get a $K$ times larger augmented training set. After that, we average the gradient of each $K$-samples to update the parameters similar in Alg. (\ref{alg:SPACE}). Results on this setting is presented in the fourth row of Table \ref{tab:defect}, where we observe a slight performance improvement on CodeBERT and a performance drop on GraphCodeBERT. This observation demonstrates that simple data augmentation is not able to enhance the model as effectively as our SPACE. Besides, training with actual augmented samples via code transformations can even hurt the performance, which is also observed in the paper of \citet{semanticroust}.

Now that actual augmented samples can not help, do the improvements just come from virtual augmented samples with any kind of perturbations on the embedding space? To answer this question, we add random perturbations on all tokens (denoted as \emph{+rand. ADV.} in Table \ref{tab:defect}) or with the same semantic-preserving way as SPACE (denoted as \emph{+rand. SPACE}) to see their effect on baseline models. Note that we also perform random perturbations $K$ times and average the gradient of the obtained $K$ samples, for a fair comparison. As shown in the last two rows of Table \ref{tab:defect}, random perturbations are sub-optimal compared with those by solving the inner maximization problem, resulting in just slight performance gains on both models. In addition, we also observe that random perturbations added with the semantic-preserving way are constantly better than those on all tokens, which further confirms the motivation of our SPACE. In conclusion, gradient-based perturbations working in tandem with the semantic-preserving mechanism (i.e., SPACE) result in the best performance.

\subsection{Robustness against Advanced Attacks}
\label{sec:robusttest}
\begin{table}[tbp]
    \centering
    \small
    \begin{tabular}{l c c c}
        \specialrule{0.09em}{0.0pt}{1.8pt}
        \multirow{2}{*}{\textbf{Model}} & \multicolumn{3}{c} {\textbf{Attack Success Rate}} \\
        & \small{MHM} & \small{Greedy-A} & \small{ALERT}\\
        \specialrule{0.03em}{1.3pt}{0.5pt}
        \specialrule{0.03em}{0.5pt}{1.3pt}
        baseline & 55.17 & 71.89 & 76.95 \\
        \quad +augmentation & 56.24 & 69.91 & 75.91 \\
        \quad +rand. ADV. & 54.89 & 70.69 & 74.75 \\
        \quad +ADV. & 52.36 & 69.34 & 73.53 \\
        \quad +rand. SPACE & 50.27 & 65.81 & 69.54 \\
        \quad +SPACE & \textbf{44.32} & \textbf{61.01} & \textbf{65.89} \\
        \specialrule{0.09em}{1.2pt}{0.0pt}
    \end{tabular}
    \caption{Results of advanced attacks on Defects4J dataset of GraphCodeBERT, where \textbf{lower ASRs are better}. Greedy-A means ALERT attack with Greedy Algorithm while ALERT stands for the one with Genetic Algorithm.}
    \label{tab:defectadv}
\end{table}

In Section \ref{sec:expresults}, we preliminarily demonstrate the robustness of SPACE in the Python adversarial test set provided by \citet{codexglue}. To test SPACE with stronger adversarial attacks, we conduct comprehensive experiments with various kinds of source code attacks, including state-of-the-art ones.

We first introduce two advanced attacking methods we experiment with. \citet{advref1aaai} propose Metropolis-Hastings Modifier (MHM) to attack code classification models, which defines identifier renaming as a sampling problem. MHM is a classical Markov chain Monte Carlo sampling approach, which generates a sequence of adversarial examples given the current stationary distribution and the probability output by the classification model. \citet{natural-attack} present ALERT (Naturalness Aware Attack) that considers the natural semantic of generated adversarial examples. They make use of the masked language prediction function of PrLMs to produce a ranked list of potential substitutes for each token, and design genetic algorithms to search for the optimal combination of variables and corresponding substitutes. ALERT is the current state-of-the-art attacking method that yields a high Attack Success Rate (ASR) on strong source code comprehension models. Meanwhile, they are the first to attack PrLMs for code.

\begin{table*}[tb]
    \centering
    \small
    \begin{tabular}{l|c c c c c c c}
		\specialrule{0.09em}{0.0pt}{0.1pt}
		\multirow{2}{*}{Model} &\multicolumn{7}{c}{CodeSearch-Adv}\\
		\cline{2-8}
		& Ruby & JavaScript & Go & Python & Java & PHP & Average \\
		\hline
		CodeBERT & 62.6($\downarrow$ 5.3) & 54.8($\downarrow$ 7.2) & 84.1($\downarrow$ 4.1) & 61.1($\downarrow$ 6.1) & 62.1($\downarrow$ 5.5) & 58.2($\downarrow$ 4.6) & 63.8($\downarrow$ 5.5)\\
		\quad +ADV. & 63.3($\downarrow$ 5.2) & 55.5($\downarrow$ 6.8) & 85.6($\downarrow$ 2.9) & 62.1($\downarrow$ 5.4) & 62.9($\downarrow$ 4.7) & 58.7($\downarrow$ 3.9) & 64.7($\downarrow$ 4.8)\\
		\quad +SPACE & \textbf{67.1}($\bm{\downarrow}$ \textbf{2.3}) & \textbf{59.5}($\bm{\downarrow}$ \textbf{3.2}) & \textbf{88.5}($\bm{\downarrow}$ \textbf{0.6}) & \textbf{65.8}($\bm{\downarrow}$ \textbf{2.5}) & \textbf{66.5}($\bm{\downarrow}$ \textbf{2.0}) & \textbf{62.3}($\bm{\downarrow}$ \textbf{1.0}) & \textbf{68.3}($\bm{\downarrow}$ \textbf{1.9})\\
		\specialrule{0.03em}{0.0pt}{0.8pt}
        \specialrule{0.03em}{0.8pt}{0.0pt}
		GraphCodeBERT & 63.2($\downarrow$ 7.1) & 57.2($\downarrow$ 7.2) & 84.5($\downarrow$ 5.2) & 62.5($\downarrow$ 6.7) & 62.6($\downarrow$ 6.5) & 58.2($\downarrow$ 6.7) & 64.7($\downarrow$ 6.6)\\
		\quad +ADV. & 65.3($\downarrow$ 5.4) & 59.1($\downarrow$ 5.8) & 86.9($\downarrow$ 2.8) & 64.9($\downarrow$ 4.6) & 64.8($\downarrow$ 4.6) & 61.3($\downarrow$ 3.4) & 67.1($\downarrow$ 4.4)\\
		\quad +SPACE & \textbf{69.8}($\bm{\downarrow}$ \textbf{3.2}) & \textbf{62.9}($\bm{\downarrow}$ \textbf{2.3}) & \textbf{89.2}($\bm{\downarrow}$ \textbf{2.9}) & \textbf{67.8}($\bm{\downarrow}$ \textbf{0.9}) & \textbf{68.1}($\bm{\downarrow}$ \textbf{2.2}) & \textbf{64.1}($\bm{\downarrow}$ \textbf{1.3}) & \textbf{70.3}($\bm{\downarrow}$ \textbf{2.1})\\
		\specialrule{0.09em}{0.1pt}{0.0pt}
	\end{tabular}
    \caption{Adversarial results on CodeSearchNet dataset, where the downward arrows indicate the performance drop compared with the original clean test set, smaller drop means better robustness.}
    \label{tab:nlcodesearchadv}
\end{table*}

\begin{table*}
    \centering
    \small
    \begin{tabular}{l|ccccc}
        \specialrule{0.09em}{0.0pt}{0.2pt}
        \multirow{2}{*}{Model} & \multicolumn{5}{c}{Python} \\
        \cline{2-6}
        & BLEU & ROUGE & METEOR & EM & F1 \\
        \hline
        GraphCodeBERT & 32.40($\downarrow$ 3.55) & 27.95($\downarrow$ 4.36) & 10.78($\downarrow$ 2.61) & 4.21($\downarrow$ 1.65) & 28.64($\downarrow$ 4.88)\\
        \quad +ADV. & 33.32($\downarrow$ 3.25) & 28.94($\downarrow$ 4.35) & 11.69($\downarrow$ 2.46) & 4.74($\downarrow$ 1.50) & 29.98($\downarrow$ 4.54)\\
        \quad +SPACE & \textbf{34.91}($\bm{\downarrow}$ \textbf{2.12}) & \textbf{30.52}($\bm{\downarrow}$ \textbf{3.45}) & \textbf{12.37}($\bm{\downarrow}$ \textbf{2.28}) & \textbf{5.21}($\bm{\downarrow}$ \textbf{1.20}) & \textbf{31.72}($\bm{\downarrow}$ \textbf{3.53})\\
        \specialrule{0.03em}{0.0pt}{0.8pt}
        \specialrule{0.03em}{0.8pt}{0.0pt}
        \multirow{2}{*}{Model} & \multicolumn{5}{c}{Java} \\
        \cline{2-6}
        & BLEU & ROUGE & METEOR & EM & F1 \\
        \hline
        GraphCodeBERT & 30.12($\downarrow$ 3.10) & 25.28($\downarrow$ 3.96) & 8.48($\downarrow$ 2.30) & 4.96($\downarrow$ 1.56) & 26.36($\downarrow$ 3.86)\\
        \quad +ADV. & 31.21($\downarrow$ 2.63) & 26.95($\downarrow$ 3.46) & 9.35($\downarrow$ 1.98) & 5.42($\downarrow$ 1.34) & 27.88($\downarrow$ 3.54)\\
        \quad +SPACE & \textbf{32.86}($\bm{\downarrow}$ \textbf{1.25}) & \textbf{28.34}($\bm{\downarrow}$ \textbf{2.62}) & \textbf{10.26}($\bm{\downarrow}$ \textbf{1.42}) & \textbf{6.18}($\bm{\downarrow}$ \textbf{0.74}) & \textbf{29.33}($\bm{\downarrow}$ \textbf{2.65})\\
        \specialrule{0.09em}{0.2pt}{0.0pt}
    \end{tabular}
    \caption{Adversarial results on CodeQA dataset, where the downward arrows means the same in Table \ref{tab:nlcodesearchadv}.}
    \label{tab:codeqaadv}
\end{table*}

Results of MHM and ALERT attacks are tabulated in Table \ref{tab:defectadv}, where we use ASR as the evaluation metric following \citet{natural-attack}. Suppose $N^+$ represents the number of correctly predicted samples on the original clean test set, while $N^-$ stands for the number of those that become incorrect after being attacked. ASR is then calculated as $\operatorname{ASR} = \frac{N^-}{N^+}$. According to the definition, lower ASR means stronger robustness against adversarial attacks. We see from the table that our SPACE is the most robust one among all models, which reduces the ASR of GraphCodeBERT by over \textbf{10\%} on all attacking methods. The second row of the table tells us that training models with \emph{actual} augmented samples by semantic-preserving code transformations can hardly help to improve the robustness against advanced attacking methods. When it comes to training with \emph{virtual} adversarial samples, Table \ref{tab:defectadv} shows that gradient-based perturbations (\emph{+ADV., +SPACE}) are constantly better than random perturbations (\emph{+rand. ADV., +rand. SPACE}), and adding perturbations under the constraint of preserving semantics (\emph{+rand. SPACE, +SPACE}) are constantly better than simply adding them on all tokens (\emph{+rand. ADV., +ADV.}). These observations further demonstrate the significance of combining gradient-based adversarial training with the data characteristics of programming languages.

We explain the superiority of SPACE from the following two perspectives. Firstly, the gradient-based perturbations are stronger than the random ones since they are basically the worst-case ones. Training with worse adversarial examples is more likely to generate more robust models. Secondly, performing semantic-preserving adversarial perturbations is consistent with the data characteristics of programming languages, which maintains grammatical correctness and avoids label flipping problem caused by small perturbations on keywords.

\subsection{Robustness against Code Transformations}
MHM and ALERT require the probability output by classification models, thus incompatible with retrieval and generative tasks. To evaluate the robustness of SPACE on these tasks and also to test its robustness under various semantic-preserving code transformations, we select the top-three effective code transformations in the experiments of \citet{semanticroust} to construct a model-agnostic adversarial testing set for both CodeSearchNet and CodeQA. 
Note that on the CodeQA dataset, some questions and answers contain identifier names on the corresponding code snippets. Therefore, when performing the identifier-renaming transformation, we also rename those identifiers on questions and answers to ensure their correctness.

Results on the adversarial test sets of both datasets are presented in Table \ref{tab:nlcodesearchadv} and Table \ref{tab:codeqaadv}, respectively. Similar observations can be seen in both tables that SPACE is the most robust model against attacks by the combination of various semantic-preserving code transformations. On the CodeSearchNet dataset, our SPACE reduces the performance drop by \textbf{3.6\%} and \textbf{4.5\%} on average over all languages for CodeBERT and GraphCodeBERT, respectively, compared with the baseline models. We also notice that vanilla adversarial training can mitigate the impact of adversarial attacks, yet only to a slight extent (typically less than 0.5\%). We infer that this is because vanilla adversarial training on all tokens is not in line with the legal semantic-preserving code transformations, since it breaks the grammatical correctness and affects the semantic meaning of keywords, which are vital and sensitive in code texts. On the contrary, our SPACE only perturbs the embedding of identifiers and leaves the keywords unchanged, which preserves the grammatical correctness and functionality of codes.

\section{Conclusion}
In order to fill the gap of performance and robustness for source code comprehension models, we propose Semantic-Preserving Adversarial Code Embeddings (SPACE) in this paper. SPACE takes the advantage of adversarial training and is specialized to fit the data characteristics of programming languages. To the best of our knowledge, we are the first to explore the efficacy of adversarial training on the continuous embedding space for source code comprehension tasks. Thorough experiments and analysis on three kinds of tasks have justified the effectiveness and universality of our model to improve both performance and robustness for source code comprehension models.

\bibliography{anthology,custom}

\begin{table*}[tbp]
    \centering
    \begin{tabular}{l|c|c|c|c}
        \specialrule{0.09em}{0.0pt}{0.2pt}
        Language & Training Samples & Dev Queries & Testing Queries & Code Database \\
        \hline Go & 167,288 & 7,325 & 8,122 & 28,120 \\
        Java & 164,923 & 5,183 & 10,955 & 40,347 \\
        JavaScript & 58,025 & 3,885 & 3,291 & 13,981 \\
        PHP & 241,241 & 12,982 & 14,014 & 52,660 \\
        Python & 251,820 & 13,914 & 14,918 & 43,827 \\
        Ruby & 24,927 & 1,400 & 1,261 & 4,360 \\
        \specialrule{0.09em}{0.0pt}{0.1pt}
    \end{tabular}
    \caption{Statistics of CodeSearchNet Dataset.}
    \label{tab:statcodesearch}
\end{table*}

\begin{table}[tbp]
\centering
    \begin{tabular}{l|c|c}
        \specialrule{0.09em}{0.0pt}{0.2pt}
         & Java & Python\\
        \hline
        Training Samples & 95,778 & 56,085\\
        Dev Samples & 12,000 & 7,000 \\
        Testing Samples & 12,000 & 7,000 \\
        \specialrule{0.09em}{0.0pt}{0.1pt}
    \end{tabular}
    \caption{Statistics of CodeQA Dataset.}
    \label{tab:codeqastat}
\end{table}

\begin{table*}[tbp]
    \centering
    \begin{tabular}{l|cc|cc}
        \specialrule{0.09em}{0.0pt}{0.2pt}
        \multirow{2}{*}{Language} & \multicolumn{2}{c|}{CodeBERT} & \multicolumn{2}{c}{GraphCodeBERT} \\
        \cline{2-5}
        & learning rate & steps & learning rate & steps\\
        \hline
        Ruby & 5e-4 & 3 & 5e-5 & 3\\
        JavaScript & 5e-5 & 3 & 5e-5 & 3\\
        Go & 5e-5 & 3 & 5e-5 & 3\\
        Python & 5e-5 & 3 & 5e-4 & 3\\
        Java & 5e-5 & 3 & 5e-5 & 3\\
        PHP & 5e-5 & 3 & 1e-4 & 3\\
        \specialrule{0.09em}{0.2pt}{0.0pt}
    \end{tabular}
    \caption{Hyper-parameters for Adversarial Training on CodeSearchNet Dataset.}
    \label{tab:hypernlcodesearch}
\end{table*}

\newpage
\appendix
\section{Dataset Statistics}
\label{app:datasetstat}

\subsection{Defects4J Dataset}
To propose a high-quality dataset for the defect detection task, \citet{defect} invested a team of security to manually label the data. They first collect functions from 4 large C-language open-source projects: Linux Kernel, QEMU, Wireshark, and FFmpeg, then ask the team of four professional security researchers to perform a two0round data labeling and cross-verification. The final dataset contains $21,000$ samples for training, $2,700$ samples for validation and $2,700$ samples for testing.

\subsection{CodeSearchNet Dataset}
This dataset contains six programming languages, which is naturally suitable to test language-agnostic methods. In our experiments, we use the filtered dataset by \citet{GraphCodeBert}, where for each language, it contains a training set denoted as $\{code, query\}_{i=1}^N$. For validation and testing, the answer is supposed to be retrieved from a code database given the query. The detailed statistics of this dataset are tabulated in Table \ref{tab:statcodesearch}.

\subsection{CodeQA Dataset}
CodeQA contains two subsets for Java and Python, respectively. The detailed statistics of this dataset are shown in Table \ref{tab:codeqastat}.

\section{Experimental Settings}
\label{app:expsetting}
\subsection{Environments}
We conduct our experiment with Pytorch implementation and CUDA version 11.4. For Defects4J and CodeQA, we train our model on an RTX 3090 24GB GPU, and for CodeSearchNet, we train our model on a Tesla V100 32GB GPU. The default settings of our experiments require about 20GB of GPU memory. Our experiments are single-runs over the same seed due to the limits of computational resources.

\subsection{Evaluation on CodeSearchNet}
Note that we use the same settings for evaluation on the CodeSearchNet dataset following \citet{GraphCodeBert}, where during validation and testing, the answer should be retrieved from the whole code database, rather than from 1,000 candidates as in the original paper of \citet{CodeSearchNet}. We use this setting for a fair comparison with the baselines of CodeBERT and GraphCodeBERT. 

\subsection{Hyper-parameters}
We search for the best hyper-parameters on the validation set of each dataset and test the best model on the test set. We list the best hyper-parameters we use in our experiments here.
\subsubsection{Defect Detection}
For both CodeBERT and GraphCodeBERT, we set the learning rate to 2e-5, batch size to 16, and the number of epochs to 5. For the inner gradient ascent of adversarial training, we set its steps to 3 steps and its learning rate to 5e-4. For CodeBERT, the maximum sequence length is 512. While for GraphCodeBERT, which requires extra data flow information, we set the maximum sequence length for code to 384 and the maximum data flow length to 128.

\subsubsection{Natural Language code Search}
For normal hyper-parameters, we set the maximum code sequence length to 320, data flow length to 64, and maximum query length to 128. The learning rate, batch size, and the number of training epochs are set to 2e-5, 32, and 10, respectively, following the original settings of \citet{GraphCodeBert}.

For hyper-parameters of adversarial training (i.e., adversarial steps and adversarial learning rate), we tune them for each language, which is presented in Table \ref{tab:hypernlcodesearch}.

\subsubsection{Question Answering over Source Code}
For normal hyper-parameters, the learning rate, batch size, and the number of training epochs are set to 1e-4, 64, and 20, respectively. For CodeBERT, the maximum code sequence length is set to 256, while for GraphCodeBERT, the maximum code sequence length and data flow length is set to 240 and 60, respectively.

For hyper-parameters of adversarial training, we set the adversarial steps to 3 for all languages on both CodeBERT and GraphCodeBERT. As for the adversarial learning rate, we set it to 5e-5 for all languages on GraphCodeBERT, and 5e-4, 1e-4 for Java and Python on CodeBERT, respectively.

\end{document}